\documentclass[10pt, a4paper]{article}

\usepackage{tar2022}

\usepackage[utf8]{inputenc}
\usepackage[pdftex]{graphicx}
\usepackage{booktabs}
\usepackage{amsmath}
\usepackage{amssymb}

\title{Abstractive Summarization as Augmentation for Document-Level Event Detection}

\name{Janko Vidaković, Filip Karlo Došilović, Domagoj Pluščec} 

\address{
University of Zagreb, Faculty of Electrical Engineering and Computing\\
Unska 3, 10000 Zagreb, Croatia\\ 
\texttt{janko.vidakovic@fer.hr}
}

\abstract{
Transformer-based models have consistently produced substantial performance gains across a variety of NLP tasks, compared to shallow models. However, deep models are orders of magnitude more computationally expensive than shallow models, especially on tasks with large sequence lengths, such as document-level event detection.
In this work, we attempt to bridge the performance gap between shallow and deep models on document-level event detection by using abstractive text summarization as an augmentation method.
We augment the DocEE dataset by generating abstractive summaries of examples from low-resource classes. 
For classification, we use linear SVM with TF-IDF representations and RoBERTa-base.
We use BART for zero-shot abstractive summarization, making our augmentation setup less resource-intensive compared to supervised fine-tuning.
We experiment with four decoding methods for text generation, namely beam search, top-k sampling, top-p sampling, and contrastive search.
Furthermore, we investigate the impact of using document titles as additional input for classification.
Our results show that using the document title offers $2.04\%$ and $3.19\%$ absolute improvement in macro F1-score for linear SVM and RoBERTa, respectively.
Augmentation via summarization further improves the performance of linear SVM by about $0.5\%$, varying slightly across decoding methods. 
Overall, our augmentation setup yields insufficient improvements for linear SVM compared to RoBERTa.
}

\begin{document}

\maketitleabstract

\section{Introduction}
Research in document-level event detection is hindered by the lack of high-quality and large-scale datasets \citep{tong2022docee}.
Even existing datasets, such as DocEE, suffer from data-related problems, such as class imbalance. 
Due to the prohibitive cost of creating new datasets, improving performance without additional labeling, for example with data augmentation, is highly desirable. Data augmentation has proved useful across various modalities, including vision and language. Therefore, using data augmentation for document-level event extraction seems like a promising research direction.

Augmentation techniques are often framed as label-preserving transformations, meaning that the label is considered unchanged after the example had gone through augmentation. 
However, the complexity of performing label-preserving transformations depends greatly on data modality. 
For example, data augmentation in a continuous space, such as pixel space in computer vision, is relatively simple to perform and often brings substantial performance benefits \citep{krizhevsky2017imagenet}. 
Conversely, augmenting data in the discrete space of natural language tokens is much less trivial.

Successful augmentation techniques in NLP often leverage deep models for text generation. Generative approaches include back-translation \citep{sennrich2016improving} and summarization \citep{yang2019data}.
However, the quality of generated text depends greatly on the chosen decoding method, i.e. beam search.
Furthermore, generation-based augmentation can become prohibitively expensive, especially when the downstream task is also tackled using a deep transformer-based model. 
On the other hand, solving downstream tasks with shallow models is undesirable because of the large performance gap between shallow and deep models, even though shallow models are computationally orders of magnitude cheaper.

In this work, we use abstractive summarization as a data augmentation technique for document-level event detection. 
We hypothesize that the performance can most easily be improved on classes comprising relatively few examples. 
We use the term "low-resource classes" when referring to such classes, and we augment only examples from those classes.
We generate abstractive summaries using documents from low-resource classes as inputs to BART generative model. We experiment with four different decoding methods for text generation, namely beam search, top-k sampling, top-p sampling and contrastive search.
We study the impact of our augmentation scheme on the performance of a shallow model, namely linear SVM. We compare this performance to a deep baseline, namely RoBERTa for sequence classification. 

Our results show that all decoding methods bring a slight performance improvement for SVM when abstractive summarization is used as an augmentation technique. Furthermore, there appears to be no significant difference between decoding methods.
We additionally experiment with using title as additional model input, both during abstractive summarization and event classification.
We find that using the document title as additional input improves the model performance. Finally, we find that even with abstractive summarization used as augmentation, the performance of SVM still lags significantly behind the performance of RoBERTa.

\section{Related Work}
Existing research on event extraction in NLP can be classified into two paradigms.
On the one hand, sentence-level event extraction deals with extracting events from singular sentences. Various event extraction tasks, such as trigger detection, trigger classification, argument identification, and argument role classification \citep{li2021compact}, are solved using datasets such as MAVEN \citep{wang2020maven} and ACE \citep{walker2006ace}.  
% TODO - add some concrete models (or results)
% TODO - explain why this is both good and bad
On the other hand, document-level event extraction aims to extract event information not from sentences, but from whole documents, such as news or Wikipedia articles. 
While document-level event detection is arguably more useful for practical applications, the research in this area is still severely limited. 
\citet{tong2022docee} identify the scarcity of large-scale datasets as one of the main reasons for this lack of research in document-level event detection. Consequently, they introduce DocEE, a large-scale dataset for document-level event detection. To the best of our knowledge, we are the first to use data augmentation on DocEE.

Notable generative methods for text augmentation include back-translation, but also summarization. In back-translation, an exemplary text is translated from the source language (i.e. English) into some target language (i.e. German) and then back-translated into source language \citep{sennrich2016improving}.
Abstractive text summarization had also been used for text augmentation \citep{yang2019data}, although not for document-level event detection.
Not only do these generative approaches depend on the availability of high-quality text generation models, but the quality of the generated text is highly sensitive to the choice of the decoding method. Beam search is the most well-established method for decoding language model outputs into text \citep{beam}. More recently developed decoding methods include top-k sampling \citep{topk}, top-p (nucleus) sampling \citep{topp}, and contrastive search \citep{sucontrastive}. \citet{su2022contrastive} show that the performance of decoding methods varies across tasks, and no method is universally the best. As a consequence, we test a multitude of decoding methods in our setup.

\section{Method}
\subsection{Task Description}

We focus on the task of document-level event detection.
In the context of machine learning, document-level event detection is a multiclass classification problem. 
Each document is assigned one class label, with classes representing event types, and documents being a textual description of the event.
We use two different models to solve the task, namely linear SVM as a shallow model and a transformer-based encoder as a deep model. 
We evaluate models using the macro F1 score, as we find all classes to be equally important.

\subsection{Dataset}

\begin{figure}
\begin{center}
\includegraphics[width=\columnwidth]{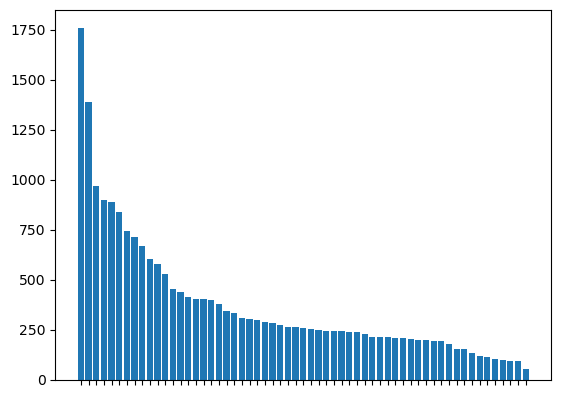}
\caption{Amount of examples per class in DocEE training set.}
\label{fig:class-freq}
\end{center}
\end{figure}

DocEE is a large-scale dataset for document-level event detection \citep{tong2022docee}.
The dataset contains $27485$ examples, and each example contains the title, the main text, and the class label which represents the event type.
The train and test sets comprise $21949$ and $5526$ examples, respectively.
A random dataset example is shown in Table~\ref{tab:example}.
There are a total of 59 event types present in the dataset. 
As can be seen in Figure~\ref{fig:class-freq}, the class imbalance is certainly present.
The most numerous class comprises $1760$ examples, and only 12 classes contain more than $500$ examples. 
We refer to classes containing less than $500$ examples as low-resource classes. In total, $11285$ examples belong to low-resource classes. We augment only examples from low-resource classes, as we hypothesize that such a setup would bring the greatest performance improvements.

\begin{table*}[ht]
\caption{A single example from the DocEE dataset. The document text is truncated.}
\centering
\label{tab:example}
\begin{tabular}{|p{5cm}|p{8cm}|c|}
\hline
\textbf{Title} & \textbf{Text} & \textbf{Event Type} \\
\hline
Covid Cost 4 Times More Jobs Than 2009 Financial Crisis: UN & Unemployment in Taiwan surged in June as companies in the service sector and their workers bore the brunt of the government measures aimed at slowing the spread of the Covid-19 outbreak. The unemp... & Financial Crisis \\
\hline
\end{tabular}
\end{table*}

\subsection{Classifiers}

We employ two models for document-level event detection.
Firstly, we use a linear SVM classifier with TF-IDF document representations as inputs.
We compute TF-IDF vectors using an n-gram range, where $n \in \{1, 2, 3\}$. 
We set the minimum document frequency to $3$, and the maximum document frequency to a proportion of $0.9$.
Otherwise, we use the default hyperparameters.~\footnote{as defined in Scikit-Learn v. $1.1.2$}
Secondly, we use RoBERTa-base \citep{liu2019roberta} as the deep baseline. 
We train the deep model for 5 epochs with a batch size of 64. We use AdamW optimizer \citep{adamw} and set the learning rate to a constant of $2 \times 10^{-5}$.
We implement our experiments in SpaCy \citep{spacy}, Scikit-Learn \citep{scikit-learn}, and Huggingface \citep{wolf2020transformers}.

\subsection{Using document title as additional input}
\label{sec:title}

We observe that document titles often provide additional context about the main event type that is present in the document.
To verify this observation, we include the document title in the training examples.
We combine the title with the document text by prepending the title to the text and including a dot as a separator, to frame the title as the first sentence of the document. 
We train both baselines with and without using the title and compare the performance on the validation set.
As seen in Table~\ref{tab:title}, using titles improves the performance of both baseline models. Consequently, we choose to always use document titles as additional input in the rest of our experiments.

\begin{table}
\caption{Macro-f1 scores of baseline models on the validation set. $D$ refers to a training setup where only document text is used for model training. In $T+D$, the document title is included in model input, as described in Section~\ref{sec:title}. For SVM results, we report the mean and standard deviation across 5 cross-validation runs.}
\label{tab:title}
\begin{center}
\begin{tabular}{lllr}
\toprule
& $D$ & $T+D$ & $\Delta~F1_M$ \\
\midrule
SVM &  $80.57_{\pm0.43}$ & $82.61_{\pm0.42}$ & $2.04$\\
\midrule
RoBERTa &  $85.82$ & $89.01$ & $3.19$ \\
\bottomrule
\end{tabular}
\end{center}
\end{table}

\subsection{Summarization as augmentation}

We use abstractive summarization as an augmentation technique, which means that we generate abstractive summaries from the existing data examples and use those summaries as novel examples.
Our approach is based on the assumption that abstractive text summarization is a label-preserving transformation with respect to the main event type. 
In other words, an abstractive summary of a document retains the same class label, i.e. the main event type, as the original document. 
We use BART \citep{lewis2020bart} to generate zero-shot summaries. A zero-shot setup is necessary because there are no target summaries for DocEE articles. The absence of target summaries makes our augmentation setup effectively unsupervised. To this end, we aim to introduce an inductive bias which is beneficial for our setup. We accomplish this feature by using a version of BART which was fine-tuned on CNN/DailyMail summarization dataset \citep{nallapati2016abstractive}. In this way, we bias the model both towards the domain of news articles and towards the task of abstractive summarization.

We use each low-resource example to generate three distinct summaries, and we apply four decoding methods for text generation. Firstly, we experiment with beam search, in which we try out $3$, $5$, and $10$ beams. Secondly, we use top-k sampling, where we set $k=640$, in accordance with \citet{topk}. Thirdly, we use top-p sampling, and we set $p=0.95$ \citep{topp}. Finally, we use contrastive search, where we set $\alpha=0.6$ \citep{sucontrastive}.

\section{Results and discussion}
Augmentation results are presented in Table~\ref{tab:results}. All decoding methods offer a slight performance boost compared to no augmentation usage. Additionally, no significant difference is apparent across different decoding methods. This finding is interesting from a practical perspective, as top-p sampling and top-k sampling offer almost an order-of-magnitude reduction in the execution time when compared to beam search. Still, not a single decoding method comes close to the baseline performance of RoBERTa without augmentation. This result brings into question the validity of using a shallow model, even though a deep model is used for augmentation. 

From a practical perspective, the combination of deep augmentation and shallow classification creates a slight but scalable improvement in the execution time. On a machine with a 32-core CPU and Nvidia RTX 3090 GPU, generating $33855$ summaries using contrastive search takes 32 minutes. On the other hand, fine-tuning RoBERTa-base on DocEE can be done in 57 minutes for 5 epochs, while performing 5-fold cross-validation with linear SVM lasts for 2 minutes. Our setup is not only about twice as fast compared to fine-tuning RoBERTa but also makes it significantly more approachable to train the model multiple times, for example for hyperparameter optimization.

\begin{table}
\caption{SVM performance on different augmentation setups. ``no AUG'' refers to baseline performance, without augmentation. "AUG-x" refers to using the decoding method x for generating summaries. $nB$ stands for beam search with $n$ beams. K stands for top-k sampling. P stands for top-P sampling. C stands for contrastive decoding. For SVM results, we report the mean and standard deviation across 5 cross-validation runs.}
\label{tab:results}
\begin{center}
\begin{tabular}{lrr}
\toprule
& \text{SVM} & \text{RoBERTa} \\
\midrule
\text{no AUG} & $82.61_{\pm0.42}$ & 89.01 \\
\midrule
\text{AUG-3B} & $83.20_{\pm 0.35}$ & \\
\text{AUG-5B} & $83.13_{\pm 0.37}$ & \\
\text{AUG-10B} & $83.15_{\pm 0.44}$ & \\
\text{AUG-K} & $83.21_{\pm 0.39}$ & \\
\text{AUG-P} & $83.08_{\pm 0.43}$ & \\
\textbf{\text{AUG-C}} & $\mathbf{83.26_{\pm 0.38}}$ & \\
\bottomrule
\end{tabular}
\end{center}
\end{table}

\section{Conclusions and future work}
In this work, we show that using abstractive summarization as augmentation brings modest performance improvements for SVM.
We also demonstrate no significant differences across decoding methods, motivating the future usage of less computationally complex methods, such as top-p sampling. Moreover, we leverage the document title to additionally improve performance. Finally, we show that the performance of SVM still lags behind the performance of RoBERTa.

Future work could explore the impact of hyperparameters on the performance of each decoding method since we used the recommended defaults for hyperparameter values. Furthermore, our approach to augmentation is fully unsupervised, since our setup utilizes no information about the event type or the event arguments. Future work could explore using supervised information to perform guided summarization. Next, we use augmentation only for shallow models for efficiency, but deep models might benefit more from high-level augmentation methods such as abstractive summarization.
Finally, we assume that augmentation is most promising when performed on low-resource classes, but we test our assumption only empirically. We leave a more formal analysis of this assumption for future work.

\bibliographystyle{tar2022}
\bibliography{tar2022} 

\end{document}